\documentclass[ao]{my_iosart2x}

\pdfoutput=1 

\usepackage{graphicx}          % standard LaTeX graphics tool
\usepackage{url}
\usepackage{color}
\usepackage{xspace}
\usepackage{amssymb}  
\newcommand{\bflist}{\begin{list}{}{\setlength{\labelsep}{4mm}\setlength{\leftmargin}{9mm}\setlength{\topsep}{2mm}\setlength{\parsep}{0mm}}}
\newcommand{\eflist}{\end{list}}

\newcommand{\bmlist}{\begin{list}{}{\setlength{\labelsep}{1mm}\setlength{\leftmargin}{6mm}\setlength{\topsep}{0mm}\setlength{\parsep}{0mm}}}
\newcommand{\emlist}{\end{list}}

\newcommand{\axlabel}{a}
\newcommand{\dflabel}{d}

\newcounter{cntax}
\newcommand{\myax}[1]{\refstepcounter{cntax}\begin{small}{\bf \axlabel\thecntax\label{#1}}\end{small}}
\newcounter{cntdef}
\newcommand{\mydf}[1]{\refstepcounter{cntdef}\begin{small}{\bf \dflabel\thecntdef\label{#1}}\end{small}}
\newcounter{cntth}

\newcounter{cntfor}

\newcounter{cntaxdolce}

\newcounter{cntdfdolce}

\newcounter{cntthdolce}

\newcounter{cntdmax}

\newcounter{cntmc}

\newcommand{\refax}[1]{(\axlabel\ref{#1})}
\newcommand{\refdf}[1]{(\dflabel\ref{#1})}

\newcommand{\dolce}{\textsc{dolce}\xspace}
\newcommand {\DOLCE}{\textsc{dolce}\xspace}
\newcommand {\cdolce} {\textsc{dolce-core}\xspace}
\newcommand{\dul}{\textsc{dul}\xspace}
\newcommand{\dns}{\textsc{d\&s}\xspace}

\newcommand {\PhRegion} {\textsc{pr}}
\newcommand {\TRegion} {\textsc{tr}}
\newcommand {\Concept} {\textsc{c}}
\newcommand {\Time} {\textsc{t}}
\newcommand {\Timequality} {\textsc{tq}}
\newcommand {\Physicalquality} {\textsc{pq}}
\newcommand {\Abstractquality} {\textsc{aq}}

\newcommand {\Process} {\textsc{pro}}
\newcommand {\Accomplishment} {\textsc{acc}}
\newcommand {\Matter} {\textsc{m}}

\newcommand{\AgentivePO}{\ensuremath{\textsc{apo}}}
\newcommand{\TLdcat}{\ensuremath{\textsc{tl}}}

\newcommand{\NAgentiveSO}{\ensuremath{\textsc{naso}}}
\newcommand{\Role}{\ensuremath{\textsc{rl}}}
\newcommand {\PhObject} {\textsc{pob}}
\newcommand {\SocialObject} {\textsc{sob}}

\newcommand {\Endurant} {\textsc{ed}}
\newcommand {\Perdurant} {\textsc{pd}}
\newcommand {\PhEndurant} {\textsc{ped}}

\newcommand {\quale} {\textsf{ql}}

\newcommand {\Part} {\ensuremath{\mathsf{P}}}
\newcommand {\Overlap} {\ensuremath{\mathsf{O}}}

\newcommand {\PPart} {\ensuremath{\mathsf{PP}}}
\newcommand {\Konst} {\ensuremath{\mathsf{K}}}

\newcommand {\isa} {\ensuremath{\mathsf{IS\_A}}}

\newcommand {\present} {\ensuremath{\mathsf{PRE}}}

\newcommand{\participates}{\ensuremath{\mathsf{PC}}}
\newcommand{\participatesC}{\ensuremath{\mathsf{PC_C}}}
\newcommand{\classifies}{\ensuremath{\mathsf{CF}}}

\newcommand{\PartAt}{\ensuremath{\mathsf{P}}}
\newcommand {\PPartAt} {\ensuremath{\mathsf{PP}}}

\newcommand{\OverlapAt}{\ensuremath{\mathsf{O}}}

\newcommand{\QTd}{\ensuremath{\mathsf{qt}}}
\newcommand{\SBLxXd}{\ensuremath{\mathsf{SBL}_{X}}}
\newcommand{\QLxTxEDd}{\ensuremath{\mathsf{ql}_{T, ED}}}
\newcommand{\QLxTxPDd}{\ensuremath{\mathsf{ql}_{T, PD}}}

\newcommand {\GEM} {\ensuremath{\mathsf{GEM}}}

\newcommand\caseLabel[1]{\getcaseid\label{#1}}
\makeatletter
\newcounter{casecount}
\newcommand{\getcaseid}{%
    \setcounter{casecount}{\arabic{subsection}}
    \def\@currentlabel{Case \arabic{casecount}}%
}
\makeatother

\newcommand\subcaseLabel[1]{\getsubcaseid\label{#1}}
\makeatletter
\newcounter{subcasecount}
\newcommand{\getsubcaseid}{%
    \setcounter{casecount}{\arabic{subsection}}
    \setcounter{subcasecount}{\arabic{subsubsection}}
    \def\@currentlabel{Case \arabic{casecount}.\arabic{subcasecount}}%
}
\makeatother

\pubyear{0000}
\volume{0}
\firstpage{1}
\lastpage{1}

\begin{document}

\begin{frontmatter}

\title{DOLCE: A Descriptive Ontology for Linguistic and Cognitive Engineering\thanks{This paper is a presentation of \dolce~based on \citep{MasoloBorgo2003} and  experience acquired with its application.}} \runtitle{DOLCE}

\begin{aug}
\author{\inits{S.}\fnms{Stefano} \snm{Borgo}\ead[label=e1]{stefano.borgo@cnr.it}%
\thanks{Corresponding author. \printead{e1}.}}
\author{\inits{R.}\fnms{Roberta} \snm{Ferrario}}
\author{\inits{A.}\fnms{Aldo} \snm{Gangemi}}
\author{\inits{N.}\fnms{Nicola} \snm{Guarino}}
\author{\inits{C.}\fnms{Claudio} \snm{Masolo}}
\author{\inits{D.}\fnms{Daniele} \snm{Porello}}
\author{\inits{E.M.}\fnms{Emilio M.} \snm{Sanfilippo}}
\author{\inits{L.}\fnms{Laure} \snm{Vieu}}
%\address{Laboratory for Applied Ontology, \orgname{ISTC-CNR}, Trento, \cny{Italy}.
%\printead[presep={\\}]{e1}} 
%E-mail: dolce@loa.istc.cnr.it}
\end{aug}

	\begin{abstract}
		{\DOLCE}, the first top-level (foundational) ontology to be axiomatized, has remained stable for twenty years and today is  broadly  used  in  a  variety  of  domains. {\DOLCE} is  inspired  by  cognitive  and  linguistic  considerations  and  aims  to  model  a commonsense view of reality, like the one human beings exploit in everyday life in areas as diverse as socio-technical systems, manufacturing, financial transactions and cultural heritage. {\DOLCE} clearly lists the ontological choices it is based upon, relies on philosophical principles, is richly formalized, and is built according to well-established ontological methodologies, e.g. OntoClean. Because of these features, it has inspired most of the existing top-level ontologies and has been used to develop or improve standards and public domain resources (e.g. CIDOC CRM, DBpedia and WordNet).
		Being a foundational ontology, {\DOLCE} is not directly concerned with domain knowledge. Its purpose is to provide the general categories and relations needed to give a coherent view of reality, to integrate domain knowledge, and to mediate across domains. In these 20 years {\DOLCE} has shown that applied ontologies can be stable and  that interoperability across reference and domain ontologies is a reality.
		This paper briefly introduces the ontology and shows how to use it on a few modeling cases. 

	\end{abstract}
\end{frontmatter}

%%%%%%%%%%%%%%%%%%%%%%%%%%%%%%%%%%%%%%%%%%%%%%%%%%%
%%%%%%%%%%%%%%%%%%%%%%%%%%%%%%%%%%%%%%%%%%%%%%%%%%%

%------------------------------------------------------------------
\section*{Introduction}\label{sec:intro} 
%-----------------------------------------------------------------------------------------

As a foundational ontology, \dolce\footnote{http://www.loa.istc.cnr.it/index.php/dolce/} provides general categories and relations that can be reused in different application scenarios by specializing them to the specific domains to be modeled.

In order to rely on well-established modeling principles and theoretical bases, it is a common practice for the categories and relations of foundational ontologies
to be philosophically grounded. This is one of the reasons why the ontological analysis preceding modeling is of paramount importance. A careful choice and characterization of categories and relations produces indeed ontologies that 
have higher chances of being interoperable, or at least of understanding potential obstacles to interoperability. 
In particular, when this strategy is applied to foundational ontologies, interoperability is possible also between the domain ontologies aligned to them. 

From a philosophical perspective, 
\dolce adopts a descriptive (rather than referentialist) metaphysics, as its main purpose is to make explicit already existing conceptualizations through the use of categories whose structure is influenced by natural language, the makeup of human cognition, and social practices. As a consequence, such categories are mostly situated at a mesoscopic level, and may change while scientific knowledge or social consensus evolve. Also, \dolce's domain of discourse is formed by particulars, while properties and relations are taken to be universals.

Once  the intended meaning of the terms denoting the relevant ontology categories has been analyzed, it should be expressed in a way that is as semantically transparent as possible. To this aim, \dolce is equipped with a rich axiomatization in first-order modal logic.
Such richness greatly enhances expressiveness but, on the other hand, it makes foundational ontologies non computable, due to the well-known trade-off between formal expressiveness and computability. For this reason, approximated and partial translations expressed  in application-oriented languages are often provided, as is the case for \dolce.\footnote{Given the emphasis on formal expressivity, recall that foundational ontologies 
are not directly used for applications; rather, they provide \textit{conceptual handles} to solve cases of misunderstandings due to the limitations of expressiveness of the application languages.}

%------------------------------------------------------------------

\subsection*{A bit of history of {\dolce}}\label{subsec:assumptions}
%------------------------------------------------------------------

The first comprehensive presentation of \dolce appeared in the deliverables of the WonderWeb project in the early 2000s and
in particular \citep{MasoloBorgo2003}. 
Following this work, several application-oriented, ``lite'' versions were later published, including \dolce-lite, \dolce-ultralite, and \dolce-zero  \citep{DBLP:conf/semweb/PaulheimG15}, see \citep{presutti2016dolce} for a summary, and widely used (see also Sect. \ref{sec:usage}). The present article is mainly based on the work of \cite{MasoloBorgo2003} with the addition of concepts, e.g. roles, as introduced by \cite{BorgoMasolo2009}.

The analysis underlying the formalization of \dolce leverages the techniques of ontological engineering and the study of classes' meta-properties of the OntoClean methodology, 
firstly developed in the early 2000s by 
\cite{DBLP:journals/cacm/GuarinoW02} and later revised by \cite{GuarinoWelty2009} and \cite{DBLP:conf/birthday/Guarino09}.

A later work presented by \cite{MasoloAndVieu2004} introduced social roles and concepts 
within \dolce through a reification 
pattern,
allowing in this way to introduce them as particulars into the domain of discourse.

In 2009, $\cdolce$ was introduced in \cite{BorgoMasolo2009}. The main purpose behind this work was that of simplifying the whole system, making it more usable in applications, and at the same time acceptable under different philosophical stands. Such simplification was also intended to facilitate the task of further extending the ontology.
In particular, some of the changes introduced by $\cdolce$ are: 
the adoption of the notion of concept as an ontology category,
a better explanation on how to distinguish and formalize properties, the formalization of the notion of resemblance to facilitate the use of qualities, and 
the possibility of having more quality spaces associated to the same quality. 
Further changes include 
the definition of different parthood relations depending on ontological categories, the introduction of a notion of time regularity, and a simplification concerning the most basic categories, which in \dolce were called `endurant' and `perdurant' and which become `object' and `event' in $\cdolce$ and can be distinguished based on whether they have space or time as main dimension, respectively.

Leaving aside these theoretical studies, {\dolce} has remained fixed over the years fulfilling the purpose of top-level ontologies to provide a solid and stable basis for modeling different domains, in this way ensuring interoperability of reference and domain ontologies that use {\dolce}. Through the years, \dolce has been enriched with modules to extend and specialize it. These modules facilitate the application and coherent use of the ontology. Some extensions tackle knowledge representation's specific issues, like the modeling of roles by \cite{MasoloAndVieu2004}, of
artifacts by \cite{DBLP:conf/fois/VieuBM08} and by \cite{Borgo-FGK2014AO}, and of modules by \cite{DBLP:conf/context/FerrarioP15}. Others showed a possible integration with machine learning and in particular computer vision \citep{DBLP:books/el/17/ConigliaroFHP17}. Extensions to the modeling of social \citep{Bottazzi2009Preliminaries-t,porello2013COIN,porello2014fois} and cognitive aspects \citep{FerrarioOltramari2004,Biccheri_et_al_2020} have also been proposed.
Today {\dolce} is becoming part of the ISO 21838 standard, under development, and is available also in CLIF, a syntax of Common Logic \cite{ISO24707}.\footnote{{\dolce} in CLIF,  OWL etc. can be found at \url{http://www.loa.istc.cnr.it/index.php/dolce/} together with additional papers and materials.}

\smallskip
The remaining of the paper is organized as follows: section \ref{sec:distinctions} introduces the most fundamental categories and relations of \dolce, which are axiomatized in section \ref{sec:formalization}. With the aim of enhancing understanding, section \ref{sec:examples} shows the application of \dolce's axioms to five modeling examples. Before looking at the structure of the ontology, we shall spend some words on its history.

%===========================================
\section{ Principles and structure of \dolce}\label{sec:distinctions}
%===========================================

\begin{figure}[h]
\includegraphics[scale=0.27]{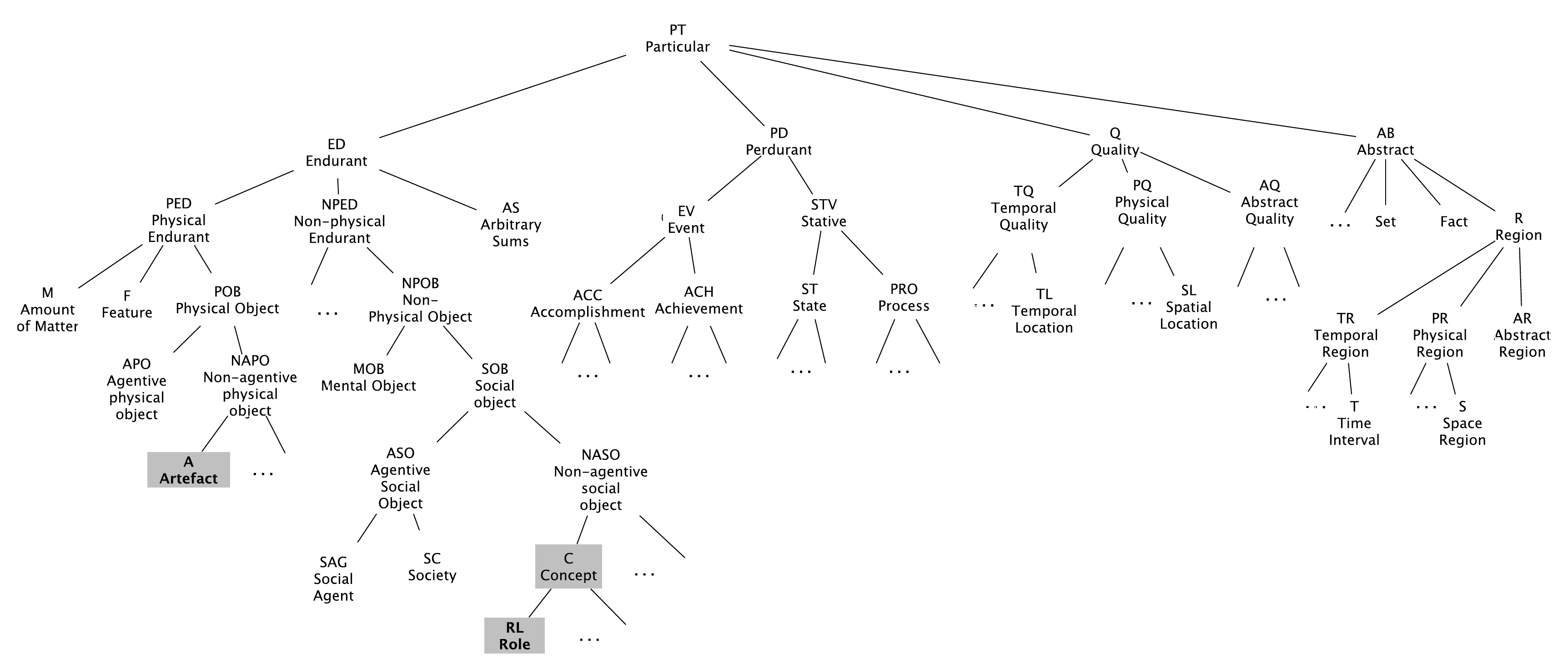}
\caption{The taxonomy of $\dolce$ extended with the subcategories \emph{Concept},  \emph{Role}, and {\em Artefact}.}
\label{fig:taxonomy}
\end{figure}

As depicted in the taxonomy in Figure \ref{fig:taxonomy}, the basic categories of \dolce are endurant (aka continuant), perdurant (occurrent), quality, and abstract.

\paragraph{I. Continuant vs. occurrent.} The distinction between endurants and perdurants is inspired by the philosophical debate about change in time. In particular, while endurants may acquire and lose properties and parts through time, perdurants are  fixed in time. Their fundamental difference concerns therefore their presence in time: endurants are wholly present (i.e., with all their parts) at any time in which they are present; differently, perdurants can be partially present, so that at any time in which they unfold only a part of them is present. 
Examples of endurants are a table, a person, a cat, or a planet, while examples of perdurants are a tennis match, a conference talk or a manufacturing process producing a certain item.

The relation connecting endurants and perdurants is called \textit{participation}. An endurant can \textit{be} in time by participating in a perdurant, and perdurants \textit{happen} in time by 
having endurants as participants. 
For instance, a person is in time by participating to her own life, and a conference talk happens if at least one  presenter (or attendant) participates to it.

\paragraph{II. Independent vs. dependent entity.} This distinction is found across the entire taxonomy of \dolce. For instance, features (e.g., edges, holes, bumps, etc.) are endurants whose existence depends on some physical object (the feature bearer), while physical objects are independent entities, i.e., their existence does not require other endurants to exist. Note that if we take a notion of cross-categorical dependence, only abstract entities turn out to be independent in {\dolce}. For instance, since a physical object necessarily participates in an event (namely, its life), every physical object requires the existence of at least one event (and vice versa).

\paragraph{III. Processes vs. events.} In {\dolce} processes and events are special types of perdurants. As it can be seen from Figure \ref{fig:taxonomy}, \dolce covers various classes of perdurant following taxonomic distinctions found in both philosophy and linguistics. In particular, a perdurant(-type) is stative or eventive according to whether it holds of the mereological sum of two of its instances, i.e. if it is cumulative or not. 
Common examples of stative perdurants are states; e.g., a sitting state is stative because the sum of two sittings is still a sitting.
Among stative perdurants, processes are cumulative but not homeomeric, namely, they have parts of different types; e.g., there are (even very short) temporal parts of a running that are not themselves runnings. Finally, eventive occurrences (events) are not cumulative, and they are called achievements if they are atomic, otherwise they are accomplishments.\footnote{As said in the Introduction, endurants are called `objects', and perdurants `events' in \cdolce. This terminological difference is due to changes in the formalization of the ontology even though the two systems largely overlap.}

\paragraph{IV. Properties, qualities, quantities.} \dolce covers these entities through the general notion of quality.\footnote{Recall that `property' is generally used in analytic metaphysics as something which can be instantiated. We treat property here in a more restricted sense; informally, as synonym of `characteristic' or `attribute'.} Qualities are, roughly speaking, what can be perceived and measured; 
they are particulars inhering in endurants or perdurants. For example, when we talk about the red of a rose, we are talking about a particular quality (that specific red) which inheres in a particular endurant (that specific rose). See also Section~\ref{sec:color}. Qualities are therefore specific to their bearers (this is why they are called {\em individual qualities} in \dolce), and they are present at each time in which their bearers are present. 
Depending on the entities in which they inhere (qualities are dependent entities indeed), \dolce identifies qualities of different types, namely, physical, temporal or abstract qualities. Moreover, since complex qualities can have qualities themselves, \dolce includes a notion of direct quality to distinguish qualities of endurants, perdurants and abstracts, from qualities of qualities.

To compare qualities of the same kind, e.g., the color of a rose and the color of a book cover, the category of \textit{quale} is introduced. A quale is the position occupied by an individual quality within a quality space.\footnote{Quality spaces in \dolce are based on G\"{a}rdenfors' conceptual spaces \citep{Gardenfors:2000vh}.} In our example, if the rose and the book cover exhibit the same shade of red, their individual colors occupy the same position (quale) in the color space. Hence, the two qualities are distinct but they have the same quale (within the same color space).

\paragraph{V. Function and Role.} \dolce does not formalize functions and roles, although these have been widely investigated and represented in \dolce-driven approaches \citep{borgo2010formalizations,MasoloAndVieu2004}. Roles are represented as
(social) concepts, which are connected to other entities (like endurants, perdurants, and abstracts) by the relation of {\em classification}. In particular, roles are concepts that are anti-rigid and founded, meaning that (i) they have dynamic properties\footnote{For instance, each role can be played by different entities at the same or at different times, the same entity can play a role at different times or discontinuously, or it can play different roles at the same or at different times.} and (ii) they have a relational nature, i.e. they depend on other roles and on contexts.

\paragraph{VI. Relations.} An important relation in \dolce is \textit{parthood}, which is time-indexed when connecting endurants and a-temporal when holding between perdurants or abstracts, i.e. between entities that do not change in time.
\textit{Constitution} is another temporalized relation in \dolce, holding between either endurants or perdurants. It is often used to single out entities that are spatio-temporally co-located but nonetheless distinguishable for their histories, persistence conditions, or relational properties. A typical example of constitution is the relation between a statue and the amount of matter it is built with. The former started to exist at a later moment with respect to the latter; the latter can survive the destruction of the former and only for the former the existence of a sculptor is a necessary condition of existence.

\smallskip
The last basic category of the ontology is that of abstracts. These are entities that have neither spatial nor temporal qualities and are not qualities themselves. We will not deal with them in the current paper, so it should suffice to give a few examples: quality regions (and therefore also quality spaces), sets, and facts. Also, although \dolce has other important categories and relations, in the present paper we will focus especially on those just presented, as they will be discussed in the following in the light of their axiomatization and used for the formalization of the cases in Section \ref{sec:examples}.

%------------------------------------------------------------------
\section{The formalization of {\dolce} in First-Order Logic}
\label{sec:formalization}
%------------------------------------------------------------------

The formal theory of $\dolce$ is written in the first-order quantified modal logic QS5, including the Barcan and the converse Barcan formula, cf. \citep{fitting2012first}.   
These assumptions entail a \emph{possibilistic} view of the entities: the domain of quantification contains all possible entities, regardless of their actual existence.

Here we present an excerpt of the axiomatization, focusing on the axioms required for the subsequent examples, that provides a general view of the $\dolce$ approach. An exhaustive presentation of $\dolce$ was given by~\cite{MasoloBorgo2003} and a proof of consistency was provided by~\cite{KutzMossakowskiAAAI2011}. 
In the following paragraphs, next to each axiom and definition we report the label of that formula in the primary presentation, cf.~\citep{MasoloBorgo2003}. 
$\dolce$ is here extended to include the category of Concepts (\textsf{C}) and Roles (\textsf{RL}) and the relation of classification (\textsf{CF}), as we shall see below; their formalization is taken from~\citep{MasoloAndVieu2004}.
\footnote{A CLIF version of DOLCE plus the theory of concepts and roles from \citep{MasoloAndVieu2004} is formalized and proved consistent by means of Mace4. The theory the proof of consistency and further material can be downloaded at \url{http://www.loa.istc.cnr.it/index.php/dolce/}} 

\subsection{Taxonomy}
As said, the taxonomy of $\dolce$ is shown in Figure \ref{fig:taxonomy}. We omit in the following the taxonomic axioms which can be found in~\citep{MasoloBorgo2003}. With respect to the original version, we include in this paper the categories \emph{Concept} and \emph{Role} as specializations of Non-Agentive Social Object, and the category \emph{Artefact} as specialization of Non-Agentive Physical Object. These will be used in the formalization of the examples.

\subsection{Mereology}

$\dolce$ assumes two primitive parthood relations: atemporal ($\Part(x,y)$ for `$x$ is part of $y$') and  time-dependent ($\Part(x,y,t)$ for `$x$ is part of $y$ at time $t$') parthood. The same predicate symbol $\PartAt$ is used for both relations. 
The first follows the principles of the General Extensional Mereology (\GEM), whereas temporary parthood drops the antisymmetry axioms, cf.~\cite[p.33]{MasoloBorgo2003}.  

Here we give some axioms and definitions relative to temporary parthood, which we will use in Section \ref{sec:artefact} (in the rest of this section Dd$n$ and Ad$n$ are the labels of definitions and axioms, respectively, used in~\citep{MasoloBorgo2003}). In the formulas, $\present(x,t)$ reads `$x$ is present at time $t$'; $\PPart(x,y,t)$ reads `$x$ is a proper part of $y$ at $t$' and $\OverlapAt(x, y, t)$ reads `$x$ and $y$ overlap at time $t$'. The expression $x +_{te} y$ reads `the temporary sum of $x$ and $y$', and $\sigma_{te} x \phi(x)$ reads `the termporary fusion of each $x$ that satisfies $\phi$'.
After the formulas we give a description in natural language. 

\bflist
\item[\myax{ax:tpart}]
$\PartAt(x, y, t) \rightarrow \Endurant(x) \wedge \Endurant(y) \wedge \Time(t)$  \hfill{}(\textit{Temporary part typing}, cf. Ad10)

\item[\myax{ax:tppresent}]
$\PartAt(x, y, t) \rightarrow \present(x,t) \wedge \present(y,t)$  \hfill{}(cf. Ad17)

\item[\mydf{df:tpp}]
$\PPartAt(x, y, t) \stackrel{def}{=} \PartAt(x, y, t) \wedge \neg \PartAt(y, x, t)$  \hfill{}(\textit{Temporary proper part}, cf. Dd20)

\item[\mydf{df:ov}]
$\OverlapAt(x, y, t) \stackrel{def}{=} \exists z(\PartAt(z, x, t) \wedge \PartAt(z, y, t))$ \hfill{}(\textit{Temporary Overlap}, cf. Dd21)

\item[\mydf{df:bsum}]
$x +_{te} y \stackrel{def}{=} \iota z\forall w,t(\OverlapAt(w, z, t) \leftrightarrow (\OverlapAt(w, x, t) \vee \OverlapAt(w, y, t)))$
\hfill{}(\textit{Temporary binary sum}, cf. Dd26)

\item[\mydf{df:usum}]
$\sigma_{te} x \phi(x) \stackrel{def}{=} \iota z\forall y,t(\OverlapAt(y, z, t) \leftrightarrow \exists w(\phi(w) \wedge \OverlapAt(y, w, t)))$\hfill{}(\textit{Temporary sum}, cf. Dd27)

\eflist 

Axiom \refax{ax:tpart} states that temporary parthood holds only between two endurants at some time, axiom \refax{ax:tppresent} states that to have a parthood relationship both the part and the whole must be present, while
\refdf{df:tpp} states that a proper part is any part which does not contain the whole itself.
\refdf{df:ov} defines overlap as a relation that holds on a pair of entities at the time when they have a common part. Using overlap, one can define
binary and unrestricted sums, see cf. \refdf{df:bsum} and \refdf{df:usum}. These definitions characterize new entities: the sum of two entities and the fusion (sum of possibly infinite entities) of all the entities that satisfy a given formula $\phi$, where $\phi$ does not contain time variables. 
Finally, note that in {\dolce} sum (fusion) is defined also on events and on abstracts, thus including the sum (fusion) of times. We do not report these latter definitions since they are standard (cf. Dd18 and Dd19). We use the same notation ($+$ and $\sigma$) for sum and fusion with or without the temporal parameter depending on the entities to which it applies.

\subsection{Quality and quale} 

The relation  \emph{being a quality of} ($\QTd$) is primitive in $\dolce$. Its full characterization is in \cite[p.35]{MasoloBorgo2003}. To be able to say that `$x$ is a quality of $y$ of type $\phi$' we extend it relatively to a type as follows:

\bflist
\item[\mydf{}]
$\QTd(\phi, x, y) \stackrel{def}{=} \QTd(x, y) \wedge \phi(x) \wedge \SBLxXd(Q, \phi)$ 
\hfill{} (\textit{Quality of type} $\phi$, cf. Dd29)
\eflist 

\noindent
where $\SBLxXd(Q,\phi)$ is an abbreviation for the statement that $\phi$ is a leaf in the \dolce hierarchy of qualities (i.e. it is a minimal category in the quality branch of Fig.\ref{fig:taxonomy}, cf. \cite[p.27]{MasoloBorgo2003}).

Then, \dolce defines the temporal \emph{quale} (relation $\textsf{ql}$), i.e., the position occupied by an individual quality within a quality space, as follows (recall that $\TLdcat$ is the temporal location category, see Fig.\ref{fig:taxonomy}): 

\bflist
\item[\mydf{TimeQualePerdurant}]	
$\QLxTxPDd(t, x) \stackrel{def}{=} \Perdurant(x) \wedge \exists z(\QTd(\TLdcat, z, x) \wedge \quale(t, z))$
\hfill{} (\textit{Temporal quale of perdurants}, cf. Dd30)

\item[\mydf{TimeQualeEndurant}]
$\QLxTxEDd(t, x) \stackrel{def}{=} \Endurant(x) \wedge t = \sigma t^\prime (\exists y(\participates(x, y, t^\prime ))$
\hfill{} (\textit{Temporal quale of endurants}, cf. Dd31)

\item[\mydf{df:quale}]
$\quale_{T}(t, x) \stackrel{def}{=} \quale_{T,ED}(t, x) \vee \quale_{T,PD}(t, x) \vee \quale_{T,Q}(t, x)$ 
\hfill{}(\textit{Temporal Quale}, cf. Dd35)
\eflist

From \refdf{TimeQualePerdurant} the temporal quale of a perdurant is the quale associated to the time location quality ($\TLdcat$) of the perdurant, and from \refdf{TimeQualeEndurant} the temporal quale of an endurant is the sum of all the times during which the endurant participates ($\participates$) to some perdurant. 
(The \emph{participation} relation is formally introduced below.) 
The temporal quale of a quality ($\quale_{T,Q}$) is defined in a similar way \cite[p.28]{MasoloBorgo2003}.
Finally, the temporal quale of an entity is given by the collection of all the previous definitions, \refdf{df:quale}.

Qualities are classified in \dolce as physical, temporal, and abstract qualities as stated below where the formulas add that a quality inheres in one and only one entity ($\QTd(x, y)$ reads `$x$ is a quality of $y$'):

\bflist
\item[\myax{PhysicalQuality}]	
$\Physicalquality(x) \rightarrow \exists !y (\QTd(x,y) \wedge  \PhEndurant(x))$
\hfill{} (\textit{{Physical quality}}, cf. Ad47)

\item[\myax{TemporalQuality}]	
$\Timequality(x) \rightarrow \exists !y (\QTd(x,y) \wedge  \Perdurant(x))$
\hfill{} (\textit{{Temporal quality}}, cf. Ad46)

\item[\myax{AbstractQuality}]	
$\Abstractquality(x) \rightarrow \exists !y (\QTd(x,y) \wedge  \textsc{nped}(x))$
\hfill{} (\textit{{Abstract quality}}, cf. Ad48)

\eflist

\subsection{Time and existence}

Actual existence in $\dolce$ is represented by means of the \emph{being present at} ($\present$) relation. The assumption here is that things exist if they have a temporal quale. 

\bflist
\item[\mydf{}]	
$\present(x, t) \stackrel{def}{=} \exists t^\prime (\quale_{T}(t^\prime , x) \wedge \PartAt(t, t^\prime ))$ \hfill{} (\textit{Being Present at} $t$, cf. Dd40) 
\eflist

Further properties of $\present$ are described in \citep{MasoloBorgo2003}, Section 4.3.8.

\subsection{Participation}\label{sec:participation}

The participation ($\participates$) relation connects endurants, perdurants, and times, i.e. endurants \emph{participate} in perdurants at a certain time \refax{pc}. Here we write $\participates(x,y,t)$ for `$x$ participates in $y$ at time $t$'. \refax{pd-pre} states that a perdurant has at least one participant and \refax{ed-pc} that an endurant participates in at least one perdurant. Axiom \refax{pc-pre} says that for an endurant to participate in a perdurant they must be present at the same time.
We also introduce the relation of constant participation ($\participatesC$), cf. \refdf{ConstantParticipation}, i.e., participation during the whole perdurant, which we will use in sections \ref{sec:EventChange} and \ref{sec:concept_evolution}.

\bflist
\item[\myax{pc}] 
$\participates(x,y,t) \rightarrow \Endurant(x) \wedge \Perdurant(y) \wedge \Time(t)$
\hfill{} (\textit{Participation typing}, cf. Ad33) 

\item[\myax{pd-pre}] 
$\Perdurant(x) \wedge \present(x,t) \rightarrow \exists y (\participates(y,x,t))$
\hfill{} (cf. Ad34) 

\item[\myax{ed-pc}] 
$\Endurant(x) \rightarrow \exists y,t (\participates(x,y,t))$
\hfill{} (cf. Ad35) 

\item[\myax{pc-pre}] 
$\participates(x,y,t) \rightarrow \present(x,t) \wedge \present(y,t)$
\hfill{} (cf. Ad36)

\item[\myax{ConstantParticipation}]	
$\participatesC(x, y) \stackrel{def}{=} \exists t(\present(y, t)) \wedge \forall t(\present(y, t) 
\rightarrow \participates(x, y, t))$ \hfill{} (\textit{Const. Participation}, cf. Dd63)

\eflist

\subsection{Constitution}

The constitution relation $\Konst$ is mainly used here to model the scenario in Section \ref{sec:artefact}. We report only a few axioms required to model the scenario ($\Konst(x,y,t)$ reads `$x$ constitutes $y$ at time $t$').

\bflist
\item[\myax{k-ed}]
$\Konst(x,y,t) \rightarrow ((\Endurant(x) \vee \Perdurant(x)) \wedge (\Endurant(y) \vee \Perdurant(y)) \wedge \Time(t))$
\hfill{} (\textit{Constitution typing}, cf. Ad20)

\item[\myax{k-ped}]
$\Konst(x,y,t) \rightarrow (\PhEndurant(x) \leftrightarrow \PhEndurant(y))$
\hfill{} (cf. Ad21)

\item[\myax{k-k}]
$\Konst(x,y,t) \rightarrow \neg \Konst(y,x,t)$
\hfill{} (cf. Ad24)

\eflist

\refax{k-ed} states that $\Konst$ applies to pairs of endurants or of perdurants and a time. \refax{k-ped} states that only physical endurants can constitute another physical endurant. \refax{k-k} states that constitution is asymmetric.

\subsection{Concepts, roles, and classification}

As anticipated, the relation of classification (\classifies) is not in~\citep{MasoloBorgo2003} as it applies to the category \emph{Concept} ($\Concept$), and to its subcategories including \emph{Role} ($\Role$), which informally collects particulars that classify, as introduced in~\citep{MasoloAndVieu2004}. We thus take the following axioms from the latter work ($\classifies(x,y,t)$ stands for `at the time $t$, $x$ is classified by the concept $y$'):

\bflist
\item[\myax{CLdom}] $\classifies(x,y,t) \rightarrow \Endurant(x) 
\wedge \Concept(y) \wedge\Time(t)$ \hfill{} (cf. A11 in \citep{MasoloAndVieu2004}\footnote{Note that \cite{MasoloAndVieu2004} apply classification only to endurants, though the possibility of applying it also to perdurants and abstracts was mentioned. Here we allow concepts to classify also perdurants as done in Section \ref{sec:EventChange}})

\item[\myax{CLpre}] $\classifies(x,y,t) \rightarrow \present(x,t)$
\hfill{} (cf. A12 in \citep{MasoloAndVieu2004})

\item[\myax{CLasym}] $\classifies(x,y,t) \rightarrow \neg \classifies(y,x,t)$
\hfill{} (cf. A14 in \citep{MasoloAndVieu2004})

\item[\myax{CLstrat}] $\classifies(x,y,t) \wedge \classifies(y,z,t) \rightarrow \neg \classifies(x,z,t)$
\hfill{} (cf. A15 in \citep{MasoloAndVieu2004})

\item[\mydf{anti-rigidity}] $\textsf{AR}(x) \stackrel{def}{=} \forall y,t (\classifies(x,y,t) \rightarrow \exists t'(\present(x,t') \wedge \neg \classifies(x,y,t'))$ 
\hfill{} (cf. D1 in \citep{MasoloAndVieu2004})

\item[\mydf{roles}] $\textsf{RL}(x) \stackrel{def}{=} \textsf{AR}(x) \land \textsf{FD}(x)$
\hfill{} (cf. D3 in \citep{MasoloAndVieu2004})
\eflist

The classification relationship $\classifies$ applies to an endurant, a concept and a time \refax{CLdom}, requires the endurant to be present when it is classified \refax{CLpre}, and is not symmetrical \refax{CLasym}. A concept can classify other concepts but not what the latter classify, this is stated to avoid circularity \refax{CLstrat}. 
Roles (\textsf{RL}) are defined as concepts that are \emph{anti-rigid} \refdf{anti-rigidity} and \emph{founded} \refdf{roles}. Informally, the foundation property (\textsf{FD}) holds for a concept that is defined by means of another concept such that the instances of the latter are all external to (not part of) the instances of the former~\citep{MasoloAndVieu2004}.

%%%%%%%%%%%%%%%%%%%%%%%%%%%%%
%%%%%%%%%%%%%%%%%%%%%%%%%%%%%
\section{Analysis and formalization in {\dolce}:  examples}\label{sec:examples}

\noindent
We present in the following sections how to formalize the five given cases according to {\dolce}.
Since to model some cases it is helpful to use a temporal ordering relation and since {\dolce} does not formalize any, we introduce one here as follows: 
`$<$' is an ordering relation over atomic and convex regions of time (usually, these are understood as time instants and time intervals) such that if $t_1 < t_2$ holds, then $t_1$ and $t_2$ are ordered and non overlapping,
i.e., $\neg \Overlap(t_1,t_2)$. 
We write $t_1 \leq t_2$ to mean that $t_1$ and $t_2$ are ordered, may properly overlap (i.e., they overlap but none is completely included in the other), and, given $t$ their overlapping region, then $t_1 - t < t_2 -t$ holds.

%%%%%%%%%%%%%%%%%%%%%%%%%%%%%

\subsection{
\caseLabel{case:const} \ref{case:const}: Composition/Constitution}\label{sec:artefact}

\begin{quote}
\begin{center}
\parbox{13cm}{``There is a four-legged table made of wood. Some time later, a leg of the table is replaced. Even later, the table is demolished so it ceases to exist although the wood is still there after the demolition.''}
\end{center}
\end{quote}

\medskip
{\dolce} provides two ways to model this and similar examples. The first option, which we call {\em artifact-based} and we follow here, considers entities like tables and legs as ontological entities on their own because of their artifactual status, namely, the fact that tables and the legs are intentionally produced products. The second option, called {\em role-based}, considers table and leg as roles of objects. In this view, indeed, some objects play the role of table and leg in a given context but not necessarily. 
We do not use this second modeling approach for \ref{case:const} and exemplify it for \ref{case:roles} (see next section) where the adoption of the role  perspective is more natural.
Note that {\dolce} is neutral with respect to the choice between these two modeling approaches: it entirely depends on what one takes as essential properties of an entity, that is, how one answers the question: is `being a table' an essential property for that object or is it only an accidental condition? In this way, by using {\dolce}, the knowledge engineer is free to choose the option that best matches their modeling purposes and application concerns.

\smallskip
Tables and legs are objects whose kinds provide  criteria for their persistence in time. We shall assume that a table remains the same object whenever it has a suitable shape and the right functionalities, even though some of its legs may be substituted. For simplicity, let us assume that a table is identified by a tabletop, i.e., no matter what happens, a table remains the same entity provided that its tabletop is not substituted or destroyed. 
Clearly, when a leg is substituted, the quantity of wood that constitutes the table changes. It follows that the existence of the table does not imply that it is made of the same matter throughout its whole life. Allowing the possibility that some entities keep existing while some of their parts change (or even cease to exist) is a design characteristic of {\dolce}. More precisely, the ontology allows distinguishing between quantities of matter (e.g., the wood of which a table is made), the object constituted by the matter (that object made of that wood), and the artifact (the table, i.e., the functional object \citep{Mizoguchi-KB2016AO}). 

The constitution and composition relations in {\dolce} capture distinct forms of dependence: the former is the dependence holding between entities with different essential properties (intercategorical) like the dependence of a table from the matter it is made of; the latter holds between entities with the same essential properties (intracategorical) like the dependence of a table from the tabletop and the legs.
It follows that constitution connects elements belonging to distinct categories and that are related by an existential co-temporal dependence. 
Here, it holds between elements of the category Matter (the considered amount of wood) and elements of the category Physical Object (the object made of that wood), since a material object exists at time $t$ only if there exists at $t$ a quantity of matter that constitutes it. 
The composition relation (expressed in {\dolce} by parthood restricted to the category at stake) holds instead among elements of the same category which are bound to form a more complex element. These are generally called composing parts or {\em components}. In this case, composition implies that the existence of the composed object requires the co-temporal existence of its composing objects. 

\smallskip
The {\dolce} categories that we use for the artifact-based modeling of this case are: matter ($\Matter$), physical object ($\PhObject$), and Time (\Time). We will also use the Artefact category, as introduced by~\cite{Borgo-V09} and two new subclasses of it introduced specifically for this scenario, i.e., Table and TableLeg.\footnote{One could avoid the use of the Artefact category and treat table and leg as mere objects. However, the introduction of domain-driven categories at intermediate layers, e.g. Artefact, is considered good practice in applications.}
In terms of relations we use: {\em being subclass} ($\isa$), {\em parthood} ($\Part$), {\em constitution} ($\Konst$) and {\em being present} ($\present$). We also use the {\em sum} operator ($+$), and the order relation ($<$) for time.

\begin{figure}[htbp]
   \centering
  \includegraphics[scale=0.48]{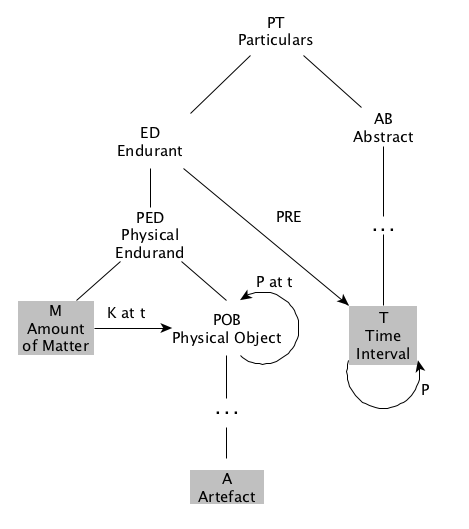} 
   \caption{Fragment of the {\dolce} taxonomy and relevant relationships for \ref{case:const}.}
   \label{fig:case_1}
\end{figure}

Figure~\ref{fig:case_1} depicts the portion of the {\dolce} taxonomy and relationships considered in this case. For the sake of simplicity, relationships like parthood ($\PartAt$) and constitution ($\Konst$) are restricted in the figure to the classes relevant for the representation of the example. Also, in all figures, ternary relations are shown in a simplified manner (e.g., $\Konst$ at $t$).

\smallskip
Formally, \ref{case:const} can be expressed as follows.

\smallskip
Taxonomic claims:
\begin{eqnarray}
\mathit{Artefact}(x) \to \PhObject(x)\\
\mathit{Table}(x) \to \mathit{Artefact}(x)\\
\mathit{Tabletop}(x) \to \mathit{Artefact}(x)\\
\mathit{Leg}(x) \to \mathit{Artefact}(x)\\
\mathit{Wood}(x) \to \Matter(x)
\end{eqnarray}

\smallskip
The previous formulas state that an artifact is a physical object, that table, tabletop and the legs are artifacts, and that wood is matter. Formula (\ref{f:table}) represents the elements and the temporal constraints ($L_{1'}$ and $W_{1'}$ are the elements which are substituted for the original table parts).

\begin{eqnarray} \label{f:table}
\nonumber
\mathit{Table}(T) \land \mathit{Tabletop}(Tp) \land \bigwedge_{1 \leq i \leq 4} \mathit{Leg}(L_i) \land \mathit{Leg}(L_{4'}) \land \mathit{Wood}(W_{top}) \land \bigwedge_{1 \leq i \leq 4} \mathit{Wood}(W_i) \land \\
\mathit{Wood}(W_{4'}) \land \Time(t) \land \Time(t') \land \Time(t'') \land t < t' \land t' < t'' 
\end{eqnarray}

The formula above states that $T$ is a table; $L_i$ are legs and so is $L_{1'}$; $W_{top}$ is an amount of wood and so are $W_i$ and $W_{1'}$ (informally, these are the amounts of wood of which the tabletop, the legs, and the new leg are made of, respectively); $t, t'$, and $t''$ are temporal instants or intervals such that $t$ is earlier than $t'$ and $t'$ is earlier than $t''$.  

\vskip10pt
Stating the elements' presence:
\begin{eqnarray} \label{f:TablePres}
\nonumber
\present(T,t) \land \present(T,t') \land 
\present(Tp,t) \land \present(Tp,t') \land 
\bigwedge_{1 \leq i \leq 4} \present(L_i,t) \land \bigwedge_{1 \leq i \leq 3} \present(L_i,t') \land \\ 
\nonumber
\present(L_{4'},t') \land \bigwedge_{1 \leq i \leq 4} \present(W_{i},t) \land \bigwedge_{1 \leq i \leq 3} \present(W_{i},t') \land \present(W_{4'},t') \land 
\bigwedge_{1 \leq i \leq 3} \present(W_{i},t'') \\ 
%\nonumber
\land \present(W_{4'},t'') \land 
\neg \present(T,t'') \land \bigwedge_{1 \leq i \leq 3} \neg \present(L_i,t'') \land \neg \present(L_{4'},t'') \phantom{abcd}
\end{eqnarray}

Formula (\ref{f:TablePres}) states that the table $T$ is present at $t$ and $t'$; the legs $L_i$ are present at $t$ and $t'$ except for $L_4$ which is not present at $t'$; $L_{4'}$ is present at $t'$; $W_{top}$ and $W_i$ are present at $t, t'$ and $t''$ except $W_{4}$ for which nothing is said about $t'$ and $t''$; $W_{4'}$ is present at $t'$ and $t''$.

\vskip10pt
Relational claims:
\begin{eqnarray} \label{f:TableRel}
\nonumber
\Part(Tp,T,t+t') \land \bigwedge_{1 \leq i \leq 4} \Part(L_i,T,t) \land \bigwedge_{1 \leq i \leq 3} \Part(L_i,T,t') \land \Part(L_{4'},T,t') \land \neg \Part(L_4,T,t') \land \\
 \Konst(W_{top},Tp,t+t') \land \bigwedge_{1 \leq i \leq 4} \Konst(W_i,L_i,t) \land \bigwedge_{1 \leq i \leq 3} \Konst(W_i,L_i,t') \land \Konst(W_{4'},L_{4'},t')
\end{eqnarray}

Formula (\ref{f:TableRel}) states that the tabletop $Tp$ is component of the table $T$ at $t$ and $t'$; the legs $L_i$ are components of $T$ at $t$; the legs $L_1, L_2, L_3$ and $L_{4'}$ are components of $T$ at $t'$; $W_{top}$ and $W_1, W_2, W_3$ are constituents of the tabletop and legs (respectively) at $t$ and $t'$; $W_{4}$ is a constituent of $L_4$ at $t$; $W_{4'}$ is a constituent of $L_{4'}$ at $t'$.

Since constitution is transitive and distributes over parthood, 
 it follows that the table $T$ is constituted by the sum of $W_{top}, W_1, W_2, W_3$ and $W_4$ at $t$, and by that of $W_{top}, W_1, W_2,  W_3$ and $W_{4'}$ at $t'$.

\medskip
The modeling presented above is mainly focused on objects: the table as a whole and the legs and tabletop as its components. In this view, the perdurants during which the table changes are not modeled. In {\dolce} one can explicitly introduce such perdurants, like the replacement and the demolition accomplishments. This second approach would make explicit the modeling of how and why the changes happen. The two views can be integrated in a single model since the essential relationships between the whole, its components and the material they are made of remain unchanged.
Other modeling views, like the functional or the role-based modelings, are also possible in {\dolce}.

%%%%%%%%%%%%%%%%%%%%%%%%%%%%%
\subsection{
\caseLabel{case:roles}\ref{case:roles}: 
Roles}

\begin{quote}
\begin{center}
\parbox{13cm}{``Mr. Potter is the teacher of class 2C at Shapism School and resigns at the beginning of the spring break. After the spring break, Mrs. Bumblebee replaces Mr. Potter as the teacher of 2C. Also, student Mary left the class at the beginning of the break and a new student, John, joins in when the break ends.''}
\end{center}
\end{quote}

\smallskip
This case requires to model social roles, thus we follow the {\em role-based} modeling approach briefly mentioned in discussing \ref{case:const}. Roles are properties that an entity can have temporarily (roles can be acquired and lost at will), and they depend on an external entity, often indicated as the context, which (perhaps implicitly) defines them. In this example, the role of student and teacher are defined within a school system, which we shall assume to stand for the context of the example.

To model \ref{case:roles}, we need four instances of Person, namely Mr. Potter, Mrs. Bumblebee, Mary, and John, as well as two instances of Object, namely, class 2C and Shapism School.\footnote{For the sake of simplicity, we ignore that a school and a class are complex objects, namely, an organization and a group. These specializations of the category Object can be modeled in \dolce by introducing the  subcategories Organization and Group following the work of \cite{Porello-BF2014Group}. Also, we do not model the `spring break' in detail and limit ourselves to see it as a generic, yet finite and temporally located, interval of time.}  

At first, say at time $t_{1}$, we have that Mr. Potter has the role of {\em teacher} (at the Shapism School's class 2C), technically writing that such role property holds for Mr. Potter at $t_{1}$. At the same time, $t_{1}$, the property does not hold for Mrs. Bumblebee. During the spring break period, say at $t_{2}$, the property holds for neither, even though the role property continues to exist, since the entities that define it (the Shapism School and the Shapism School's class 2C) continue to exist during the break. After the spring break, at $t_3$, Mrs. Bumblebee has the (Shapism School's class 2C) teacher role and Mr. Potter has not. 
The role teacher is played by a person at $t_{1}$, by nobody at $t_{2}$, and by another person at $t_{3}$. The Shapism School's class 2C teacher role exists and does not change during the whole period. 
Since the teacher role can be played by one person at a time, usually one says that Mrs. Bumblebee replaced Mr Potter in that teacher role.

Similarly, at first Mary has the {\em student} role (at the Shapism School's class 2C) and John has not. Only the persons who are students before the break and do not leave the class have the student role during the break.
Those people, now including John, have the Shapism School's class 2C student role after the break. In this case, however, one cannot say that John substituted Mary since, differently from teacher roles, which are characterized by individual rights and duties (an English teacher and a math teacher must satisfy different requirements and have duties tailored to the discipline they are hired for), the class 2C student role does not differentiate among players. 

\begin{figure}[htbp]
   \centering
    \includegraphics[scale=0.45]{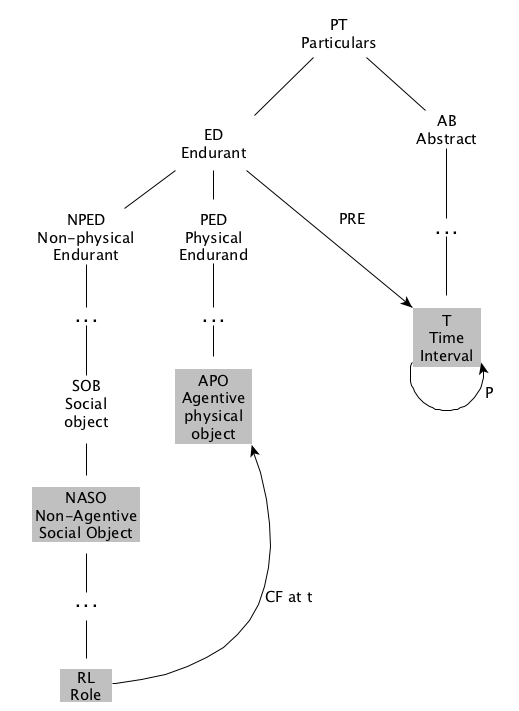} 
   \caption{Fragment of the {\dolce} taxonomy and relevant relationships for \ref{case:roles}.}
   \label{fig:case_2}
\end{figure}

\smallskip
The {\dolce} categories that we need for modeling this case are: agentive physical object (\AgentivePO), non-agentive social object (\NAgentiveSO), and Time (\Time). We will also use the Teacher and Student roles as specializations of the Role category (\Role, a subcategory of \NAgentiveSO) from~\citep{MasoloAndVieu2004}. In terms of relations we use: {\em being subclass} ($\isa$), {\em being present} ($\present$), {\em time order} ($<$), {\em mereological sum} ($+$), and the {\em classify} relation ($\classifies$) also introduced in \citep{MasoloAndVieu2004}. 
Figure \ref{fig:case_2} depicts some  relevant classes and relationships for this case. 

Formally, \ref{case:roles} can be expressed as follows.

\smallskip
Taxonomic claims:
\begin{eqnarray}
\mathit{Person}(x) \to \AgentivePO(x)\\
\mathit{Funct}\Role(x) \to \Role(x)\\
\Role(x) \to \NAgentiveSO(x)
\end{eqnarray}

The previous formulas state that a person is an agentive physical object, a functional role is a role and a role is a non-agentive social object.

\smallskip
Functional role characterization:
\begin{eqnarray} \label{f:FunRole}
\mathit{Funct}\Role(y) \land \classifies(x,y,t) \land \classifies(x',y,t) \to x=x'
\end{eqnarray}

Formula (\ref{f:FunRole}) states that a functional role ($y$) can classify only one entity at each time.

\smallskip
The elements and the temporal constraints:
\begin{eqnarray} \label{f:ConsRole}
\nonumber 
\mathit{Person}(Potter) \land \mathit{Person}(Bumblebee) \land \mathit{Person}(Mary) \land \mathit{Person}(John) \land\\
\nonumber
 \Role(2CStudent) \land \mathit{Funct}\Role(2CTeacher) \land \neg \mathit{Funct}\Role(2CStudent) \land \\
 \Time(t_1) \land \Time(t_2) \land \Time(t_3) \land t_1 < t_2 < t_3
\end{eqnarray}

Formula (\ref{f:ConsRole}) states that Potter, Bumblebee, Mary, and John are persons; that 2CTeacher and 2CStudent are roles and that the first of these is a functional role. Finally, the formula says that $t_i$ are times and indicates their ordering.

\vskip10pt
Stating the elements' presence:
\begin{eqnarray} \label{f:TemRole}
\nonumber
\present(Potter,t_1) \land \present(Bumblebee,t_2+t_3) \\
\present(Mary,t_1) \land \present(John,t_3) 
\end{eqnarray}

Formula (\ref{f:TemRole}) states that Potter, Bumblebee, Mary, and John exist at least at the listed times.

\vskip10pt
Relational claims:
\begin{eqnarray} \label{f:RelRole}
\nonumber
\forall x~ \neg \classifies(x,2CTeacher,t_2) \land \\
\nonumber
\classifies(Potter,2CTeacher,t_1) 
\land \classifies(Bumblebee,2CTeacher,t_3) \land \\
\nonumber
\classifies(Mary,2CStudent,t_1) \land \neg \classifies(John,2CStudent,t_1) \land \\
\nonumber
\neg \classifies(Mary,2CStudent,t_2) \land \neg \classifies(John,2CStudent,t_2) \land \\
\neg \classifies(Mary,2CStudent,t_3) \land \classifies(John,2CStudent,t_3) 
\end{eqnarray}

Formula (\ref{f:RelRole}) states that: 2CTeacher holds for nobody at $t_2$;
Potter satisfies 2CTeacher at $t_1$ only; Bumblebee satisfies 2CTeacher at $t_3$ only; Mary satisfies 2CStudent at $t_1$ only; John satisfies 2CStudent at $t_3$ only; neither Mary nor John satisfies  2CStudent at $t_2$.

The model presented here is the most natural approach for this kind of scenarios in {\dolce}.

%%%%%%%%%%%%%%%%%%%%%%%%%%%%%
\subsection{
Property change}
%%%%%%%%%%%%%%%%%%%%%%%%%%%%%
\subsubsection{\subcaseLabel{case:color}\ref{case:color}: color change}
\label{sec:color}

\begin{quote}
\begin{center}
\parbox{13cm}{``A flower is red in the summer. As time passes, the color changes. In autumn the flower is brown.''}
\end{center}
\end{quote}

\smallskip
We have seen how to understand and model essential properties in \ref{case:const} and roles (dynamic, contextual properties) in \ref{case:roles}. To model \ref{case:color}, we use individual qualities, that is, properties as manifested by an object. 
These are properties that an object must have, they are necessary for its existence. For instance, in the case of material objects, these include mass, color, and
speed. 
Having qualities is necessary for objects, although the value they take may change in time.

\begin{figure}[htbp]
   \centering
   \includegraphics[scale=0.5]{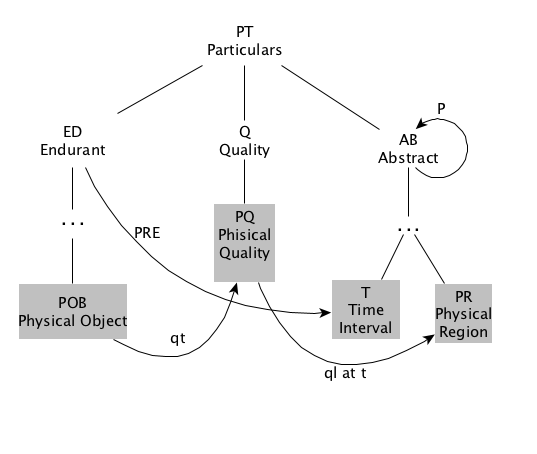} 
   \caption{Fragment of the {\dolce} taxonomy and relevant relationships for \ref{case:color}.}
   \label{fig:case_3}
\end{figure}

\smallskip
The {\dolce} categories needed to model \ref{case:color} are: physical object (\PhObject), physical quality (\Physicalquality), physical (quality) space (\PhRegion), and time (\Time). We will also use Flower as specialization of the {\PhObject} category, $\mathit{ColorQuality}$ as specialization of the {\Physicalquality} category, and $\mathit{ColorSpace}$ as specialization of the {\PhRegion} category.
For relations we use: {\em being subclass} ($\isa$), {\em inherence} ($\QTd$),  {\em being present} ($\present$), {\em parthood} ($\Part$),  {\em time order} ($<$), and (the relation) {\em quale} ($\quale$). 
Figure \ref{fig:case_3} depicts some  relevant classes and relations used for representing \ref{case:color}.

Formally, \ref{case:color} can be expressed as follows.

\smallskip
Taxonomic claims:
\begin{eqnarray}
\mathit{Flower}(x) \to \PhObject(x)\\
\mathit{ColorQuality}(x) \to \Physicalquality(x) \\
\mathit{ColorSpace}(x) \to \PhRegion(x)
\end{eqnarray}

The previous formulas state that a flower is a physical object, a color quality is a quality of physical endurants and a color space is one of the spaces in the physical region.

\smallskip
The elements we need to model this case are:
\begin{eqnarray}
\label{eq:flower}
\mathit{Flower}(F) \land \mathit{ColorQuality}(q) \land \Time(Summer) \land \Time(Autumn) \land \Time(t_0) \land \Time(t_1)
\end{eqnarray}

Formula (\ref{eq:flower}) states that $F$ is a flower, $q$ is a color quality, Summer and Autumn are times (thus, these are not modeled as seasons in this example) and so are $t_0$ and $t_1$. The following formula states that the flower $F$ is present during the Summer and the Autumn.

\smallskip
Stating the elements' presence: 
\begin{eqnarray}
\present(F,Summer)
\land \present(F,Autumn)
\end{eqnarray}

\smallskip
Relational claims: 
\begin{eqnarray} \label{f:RelColor1}
\nonumber
\QTd(q,F) \land \quale(l,q,t_0) \land
\Part(t_0,Summer) \land \quale(l',q,t_1) \land  \Part(t_1,Autumn) \land \\
\nonumber
\Part(l,RedRegion) \land 
\Part(l',BrownRegion) \land \\
\nonumber
\Part(RedRegion,ColorSpace) \land 
\Part(BrownRegion,ColorSpace)\land \\
Summer < Autumn
\end{eqnarray}

Formula (\ref{f:RelColor1}) states that: $q$ is the color quality of flower $F$; $q$ has value $l$ at time $t_0$ in the summer and has value $l'$ at time $t_1$ in the autumn where $l$ is located in the red region and $l'$ in the brown region (both regions in the color space). Finally, it states that Summer is before Autumn.

One can model that the flower takes all the shades from red to brown by adding the following formula (here $SC$ stands for the property of self-connected region, a property which is defined from the connection relation $C$ in the standard way, cf. \citep{Casati:96bm}:
\begin{eqnarray} \label{f:ChangeColor1}
\nonumber
\exists p (SC(p) \land \Part(p,ColorSpace) \land 
\Part(l,p) \land
\Part(l',p) \land \\
\forall l^* (\Part(l^*,p) \to \exists t (\Part(t,Summer+Autumn) \land \quale(l^*,q,t))))
\end{eqnarray}

Formula (\ref{f:ChangeColor1}), combined with the earlier formulas, states that there exists a path ($p$) in the space of colors which has the given red and brown colors of the flower as endpoints, and such that the flower takes all the colors in the path during the Summer and Autumn.
In a similar way, one can also model that the change of color has no jumps. For instance, preventing the flower from suddenly jumping from red to light brown, then back to scarlet etc.

The model presented here follows the approach that best exploits {\dolce}'s treatment of qualities.

%%%%%%%%%%%%%%%%%%%%%%%%%%%%%
\subsubsection{\subcaseLabel{case:speed}\ref{case:speed}: speed change}
%%%%%%%%%%%%%%%%%%%%%%%%%%%%%

\begin{quote}
\begin{center}
\parbox{13cm}{``A man is walking when suddenly he starts walking faster and then breaks into a run.''}
\end{center}
\end{quote}

This example focuses on a change that occurs during an event. The event is divided in three parts, in the first part the man is walking, that is, there is a movement based on a repeated regular movement which is a process in {\dolce}. In the second part, there is again a movement which is repeated at an increasing frequency until the desired speed is reached.\footnote{One can argue that the quality that distinguishes walking from running is not speed but how the feet touches the ground or a combination of this and the speed quality. In these cases, the modeling approach is analogous to the one we provide here, what changes is only the quality one considers.}  For this reason, we model the second part of the event as an accomplishment whose completion point is the achievement of the desired speed. Finally, the third part is a movement based on a repeated regular movement (running) which is similar to the first movement but with different characteristics. 
From this analysis, we model \ref{case:speed} as an event composed of three ordered subevents. 

\begin{figure}[htbp]
   \centering
   \includegraphics[scale=0.45]{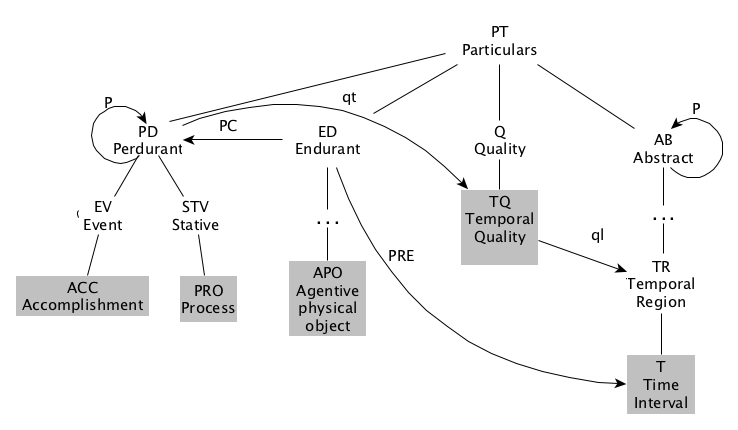} % requires the graphicx package
   \caption{Fragment of the {\dolce} taxonomy and relevant relationships for \ref{case:speed}.}
   \label{fig:case_3.2}
\end{figure}

\smallskip
The {\dolce} categories that we need for modeling \ref{case:speed} are: agentive physical object (APO), process (PRO), time quality (TQ), accomplishment (ACC), quale (\quale), and time (T). 
In terms of relations we use: {\em being subclass} ($\isa$), {\em constant participation} ($\participatesC$), {\em parthood} ($\Part$), {\em quality of} ($\QTd$), {\em being present}  ($\present$), {\em time order} ($<$), {\em mereological} sum ($+$), and (the relation) {\em quale} ($\quale$). 
Figure \ref{fig:case_3.2} depicts (some of) the classes and relationships relevant for representing \ref{case:speed}.

Formally, \ref{case:speed} can be expressed as follows.

\smallskip
Taxonomic claims:
\begin{eqnarray}
\mathit{Person}(x) \to \AgentivePO(x)\\
\mathit{SpeedQuality}(x) \to \Timequality(x)\\ 
\mathit{SpeedSpace}(x) \to \TRegion(x) \\
\mathit{Walk}(x) \to \Process(x)\\
\mathit{Run}(x) \to \Process(x)\\
\mathit{SpeedUp}(x) \to \Accomplishment(x)
\end{eqnarray}

The formulas above state that a person is an agentive physical object, speed is a quality of perdurants, a space of speed measure is a physical region, walking and running are processes, speeding up is an accomplishment.
\smallskip
The elements and the temporal constraints:

\begin{eqnarray}
\nonumber
\mathit{Person}(p) \land \Perdurant(e) \land \mathit{Walk}(e_1) \land \mathit{SpeedUp}(e_2) \land \mathit{Run}(e_3) \land \\
\nonumber
\mathit{SpeedQuality}(s) \land \mathit{SpeedQuality}(s_1) \land
\mathit{SpeedQuality}(s_2) \land
\mathit{SpeedQuality}(s_3) \land
\\
\Time(t_e) \land \Time(t_{e1}) \land \Time(t_{e2}) \land \Time(t_{e3}) \label{f:case3.2_34}
\end{eqnarray}

The formula says that $p$ is a person, that there is a perdurant $e$, a walking perdurant $e_1$, a speeding-up perdurant $e_2$, a running perdurant $e_3$, that $s$ and $s_i$ are speed qualities, and that $t_e, t_{e1}, t_{e2}, t_{e3}$ are times.

\smallskip
The following formula states that $p$ exists during the time $t_e$:
\begin{eqnarray}
\present(p,t_e)
\end{eqnarray}

\smallskip
Relational claims (note that {\dolce} already ensures that the quale "l" is in the speed space):
\begin{eqnarray}
\nonumber
\Part(l,SpeedSpace) \land \Part(l_1,SpeedSpace) \land \\
\nonumber
\Part(l_2,SpeedSpace) \land 
\Part(l_3,SpeedSpace) \land \\
\nonumber
\QTd(s,e) \land \quale(l,s,t_{e}) \land \QTd(s_1,e_1) \land \quale(l_1,s_1,t_{e1}) \land \\
\nonumber
\QTd(s_2,e_2) \land \quale(l_2,s_2,t_{e2}) \land \QTd(s_3,e_3) \land \quale(l_3,s_3,t_{e3}) \land \\ e=e_1+e_2+e_3 \land \participatesC(p,e) 
\end{eqnarray}
This formula says that $l, l_1, l_2$ and $l_3$ are locations in $SpeedSpace$. It also states that $s, s_1, s_2$ and $s_3$ are qualities of the perdurants $e, e_1, e_2$ and $e_3$, respectively, and have locations $l, l_1, l_2$ and $l_3$. Finally, it states that $p$ constantly participates in the perdurant $e$ which is the sum of the perdurants $e_1, e_2, e_3$.

\medskip
We can now characterize the core property of walking and of running: these are events  across which the speed of the participant is qualitatively stable. This is what formula (\ref{Fla:walking}) states by enforcing the speed quality of a walking (or of a running) perdurant to remain in the same position during the perdurant, say within the range for walking or for running.\footnote{In {\dolce} this can be done by measuring the quality in a qualitative speed space. For instance, take a space with two values only, say, `regular speed' and `varying speed'. When an event has only limited speed variations (e.g., according to the granularity of that space), the associated speed quale is `regular speed'.} A speeding up event is an event in which the frequency of a process increases. In the specific case, the change leads to move from a walking to a running process. To characterize events in which speed regularly changes, we introduce formula (\ref{Fla:speeding}): this formula states that there is at least one speed change during the event, and that any speed change during the event can only increase the speed (here $<_{speed}$ is the ordering in the speed quality space).
\begin{eqnarray}
\nonumber (\QTd(s,x) \land (\mathit{Walk}(x) \lor \mathit{Run}(x))) \to 
\forall l_i,l_j,t_i,t_j (\quale(l_i,s,t_i) \land \\ 
\quale(l_j,s,t_j) \land 
 \Part(t_i,t_x) \land \Part(t_j,t_x) \to l_i = l_j) 
\label{Fla:walking}
\end{eqnarray}

\begin{eqnarray}
\nonumber
qt(s,x) \land \mathit{SpeedUp}(x) \to \\
\nonumber
\exists l_i,l_j,t_i,t_j (\Part(t_i,t_{x}) \land \Part(t_j,t_{x}) \land \quale(l_i,s,t_i) \land \quale(l_j,s,t_j) \land l_i \neq l_j) \land \\
\forall l_i,l_j,t_i,t_j (\Part(t_i,t_{x}) \land \Part(t_j,t_{x}) \land \quale(l_i,s,t_i) \land \quale(l_j,s,t_j) \to (l_i \leq_{speed} l_j \leftrightarrow t_i < t_j))
\label{Fla:speeding}
\end{eqnarray}

\smallskip
{\DOLCE} and these formulas for the specific \ref{case:speed} suffice to model the example of this section. To model continuity in speed change, one can use the approach exploited in formula (\ref{f:ChangeColor1}).

As for the previous case, the model presented here shows the most natural modeling approach for this kind of scenarios in {\dolce}.

%%%%%%%%%%%%%%%%%%%%%%%%%%%%%
\subsection{
\caseLabel{case:event}\ref{case:event}: 
Event Change}\label{sec:EventChange}
%%%%%%%%%%%%%%%%%%%%%%%%%%%%%

\begin{quote}
\begin{center}
\parbox{13cm}{``A man is walking to the station, but before he gets there, he turns around and goes home.''}
\end{center}
\end{quote}

\smallskip
Following the viewpoint of {\dolce}, this case is composed of (sub)events that correspond to the execution of distinct plans: reaching the station and reaching home. The first event (a man walking to the station) and the third (a man going home) are processes that are intended to be parts of a plan execution, that is, parts of distinct accomplishments. The intermediate event is an accomplishment (turning towards a direction) which is part of the second plan, namely, reaching home. To model this case, we need to include in the formalization the purpose of the (sub) events. 

\begin{figure}[htbp]
   \centering
   \includegraphics[scale=0.45]{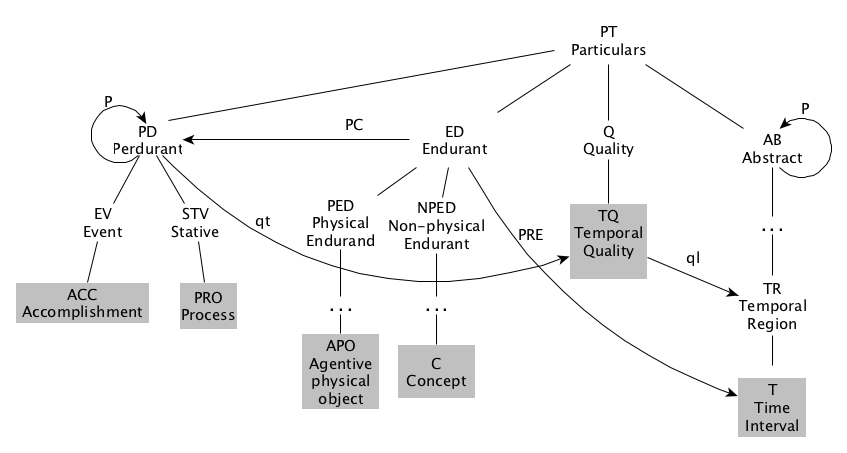}
   \caption{Fragment of the {\dolce} taxonomy and relevant relationships for \ref{case:event}}
   \label{fig:case_4}
\end{figure}

\smallskip
The {\dolce} categories that we need for modeling \ref{case:event} are: physical object (\PhObject), agentive physical object (\AgentivePO), concept (\Concept), process (\Process), accomplishment (\Accomplishment), temporal quality (\Timequality), and time (\Time). 
We will also use  $\mathit{DirectionQuality}$ and  $\mathit{SpeedQuality}$ as specialization of the quality category. 
In terms of relations we use: {\em subsumption} ($\isa$), {\em constant participation} ($\participatesC$), {\em being present} ($\present$), 
{\em mereological sum} ($+$), {\em parthood} ($\Part$), {\em quale} ($\quale$), {\em inherence} ($\QTd$), {\em classification} ($\classifies$), {\em temporal order} ($<$). In addition, we introduce the new relationship {\em ExecutesPlan} to connect a perdurant to a plan. This relation is used to state that an event complies with the plan requirements. For instance, if a plan $p$ states that a person must go first to point A and then to point B, then any event $e$ that takes that person to point A satisfies $\mathit{ExecutesPlan}(e,p)$ because it executes the plan even though it does not complete it.
Figure \ref{fig:case_4} depicts some  relevant classes and relationships.

Formally, \ref{case:event} can be expressed as follows.

\smallskip
Taxonomic claims:
\begin{eqnarray}
\mathit{Person}(x) \to \AgentivePO(x)\\
\mathit{DirectionQuality}(x) \to \Timequality(x)\\
\mathit{SpeedQuality}(x) \to \Timequality(x)\\
\mathit{Walk}(x) \to \Process(x)\\
\mathit{Turn}(x) \to \Accomplishment(x)\\
\mathit{Plan}(x) \to \Concept(x)
\end{eqnarray}

The previous formulas state that a person is an agentive physical object and that direction and speed qualities are qualities of perdurants.

The elements we need to model this case are a person, a perdurant, two walking and a turning events, two plans and three times:

\begin{eqnarray}
\nonumber
\mathit{Person}(a) \land \Perdurant(e) \land \mathit{Walk}(e_1) \land \mathit{Turn}(e_2) \land \\ 
\mathit{Walk}(e_3) \land \mathit{Plan}(p_1) \land \mathit{Plan}(p_2) \land \Time(t_{e1}) \land \Time(t_{e2}) \land \Time(t_{e3})
\end{eqnarray}

\vskip10pt
Stating the temporal constraints and the elements' presence:
\begin{eqnarray}
\nonumber
t_{e1} < t_{e2} < t_{e3} \land 
\quale_T(t_{e1},e1) \land 
\quale_T(t_{e2},e2) \land 
\quale_T(t_{e3},e3) \land 
\present(a,t_e) \land  \present(p_1,t_{e1}) \land \\ \present(p_2,t_{e2}) \land 
\present(p_2,t_{e3}) \land 
\neg \present(p_1,t_{e2}) \land \neg \present(p_1,t_{e3}) \land \neg \present(p_2,t_{e1})
\label{fla:pre4}
\end{eqnarray}

Formula (\ref{fla:pre4}) states the ordering of the times, that $t_{ei}$ is the time of perdurant $e_i$, that person $a$ is present all the times, that plan $p_1$ is present during $e_1$ and plan $p_2$ is during $e_2$ and $e_3$. It also says that plan $p_1$ is not present during $e_2$ and $e_3$ while plan $p_2$ is not present during $e_1$.

The following formula binds the use of the execution relation to pairs of one perdurant and one concept, we do not characterize it further:
\begin{eqnarray}\label{f:ExPl}
\mathit{ExecutesPlan}(x,y) \rightarrow \Perdurant(x) \land \Concept(y)
\end{eqnarray}

We now write $t_{2i}$ and $t_{2f}$ for the initial and final time of event $e_2$:
\begin{eqnarray}\label{f:change}
\nonumber
\mathit{DirectionQuality}(s) \land \QTd(s,e) \land \quale(l_{1},s,t_{e1}) \land 
\quale(l_{2},s,t_{e2}) \land \quale(l_{3},s,t_{e3}) \land \\ 
\nonumber
\quale(l_1,s,t_{2i}) \land \quale(l_3,s,t_{2f}) \land l_1 \neq l_3 \land e=e_1+e_2+e_3 \land \participatesC(a,e) \land \\
\mathit{ExecutesPlan}(e_1,p_1) \land \mathit{ExecutesPlan}(e_2+e_3,p_2)
\end{eqnarray}

Formula (\ref{f:change}) states that the direction quality $s$ of the event $e$ changes during the turning subevent $e_2$, and that event $e_1$ executes plan $p_1$ and event $e_2 + e_3$ executes plan $p_2$. Finally, it states that $e_1, e_2$ and $e_3$ span the whole event $e$ and that person $a$ participates to the whole event.

To state that an event $x$ is a walking event, we can use a formula similar to the one introduced in \ref{case:speed}, reported below as (\ref{Fla:walking2}). 
To characterize the core property of a turning event $y$, we use formula (\ref{Fla:turning}) where $l_1$ and $l_3$ are as in formula (\ref{f:change}) and write $t_y$,  $t_yi$ and $t_yf$ for the temporal interval of event $y$ and for its initial and final instants, respectively.\footnote{For completeness, one should add the symmetric condition for $l_i > l_j$.} 
\begin{eqnarray}
\nonumber \mathit{SpeedQuality}(s) \land \QTd(s,x) \land \mathit{Walk}(x) \to \\
\forall l_i,l_j,t_i,t_j (\quale(l_i,s,t_i) \land 
\quale(l_j,s,t_j) \land 
 \Part(t_i,t_x) \land \Part(t_j,t_x) \to l_i = l_j) 
\label{Fla:walking2}
\end{eqnarray}

\begin{eqnarray}
\nonumber
\mathit{DirectionQuality}(s) \land \QTd(s,y) \land \mathit{Turn}(y) \land
\quale(l_1,s,t_{yi}) \land \quale(l_3,s,t_{yf}) \land t_i < t_j \land \\
%\forall l_i,l_j,r_i,r_j,t_i,t_j (
\nonumber
l_1 < l_3 \land \Part(t_i,t_{y}) \land \Part(t_j,t_{y}) \land \quale(l_i,s,t_i) \land \quale(l_j,s,t_j) \land l_i+r_i = l_j+r_j = l_3 \to\\ 
0 \leq r_j < r_i
\label{Fla:turning}
\end{eqnarray}

The modeling approach we followed here is the preferred one in {\dolce} for this kind of scenarios.

%%%%%%%%%%%%%%%%%%%%%%%%%%%%%
\subsection{
\caseLabel{case:evolution}\ref{case:evolution}: 
Concept Evolution}\label{sec:concept_evolution}
%%%%%%%%%%%%%%%%%%%%%%%%%%%%%

\begin{quote}
\begin{center}
\parbox{13cm}{Background: marriage is a contract between two people that is present in most social and cultural systems and it can change in major (e. g. gender constraints) and minor (e.g. marriage breaking procedures) aspects. ``Marriage is a contract that is regulated by civil and social constraints. These constraints can change but the meaning of marriage continues over time.''}
\end{center}
\end{quote}

\smallskip
There is disagreement about the nature of concepts, including whether concepts can change in time while preserving identity. Some argue that concepts have a stable nature (their characterizations cannot change in time), others argue the opposite \citep{masolo2019modeling}. 
Similarly to the case of artifacts presented in Sect. \ref{sec:artefact}, {\DOLCE} does not prescribe the adoption of one or the other view, allowing in this way the knowledge engineer to select the approach that better fits with their modeling needs and world-view. For instance, the example mentioned above assumes that concepts can persist through time while partially changing in their characterization. 
In particular, it points to a  social scenario where the concepts characterizing a socio-cultural system are associated with different rules across time because of the legal and cultural evolution of the society. We shall therefore take this perspective for the sake of this case.

\begin{figure}[htbp]
   \centering
   \includegraphics[scale=0.5]{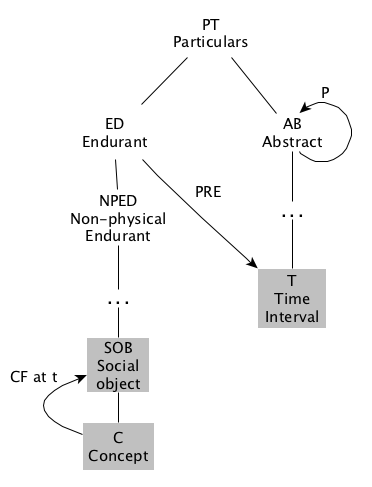} 
   \caption{Fragment of the {\dolce} taxonomy and relevant relationships for \ref{case:evolution}.}
   \label{fig:case_5}
\end{figure}

\smallskip
The {\dolce} categories that we need for modeling \ref{case:evolution} are: 
social object (\SocialObject), concept (\Concept), and time (\Time). 
In terms of relations, we use: {\em subsumption} ($\isa$), {\em being present} (\present), and {\em classification} (\classifies). 
Figure \ref{fig:case_5} depicts the {\dolce} classes and relationships used for \ref{case:evolution}.

Formally, \ref{case:evolution} can be expressed as follows.

\smallskip
Taxonomic claims (a social relationship, $\mathit{SocRelationship}$, holds for various types of unions between people; the notions of social marriage and legal marriage are intended to be elements in the {\dolce} category of concepts): 
\begin{eqnarray}
\mathit{SocMarriage}(x) \to \Concept(x)\\
\mathit{LegMarriage}(x) \to \Concept(x)\\
\mathit{SocRelationship}(x) \to \SocialObject(x)
\end{eqnarray}

\smallskip
The elements and the temporal constraints that we need are: a social relationship $M$, a social concept of marriage $sm$,  two legal concepts of marriage and two times:
\begin{eqnarray}
\nonumber
\mathit{SocRelationship}(M) \land \mathit{SocMarriage}(sm) \land \mathit{LegMarriage}(lm) \land \mathit{LegMarriage}(lm') \\ 
\land \Time(t) \land \Time (t')
\end{eqnarray}

\smallskip
The social relationship holds in both times and so does the social marriage, one legal marriage concept exists at $t$, the other at $t'$. Then, the elements' presence is as follows:
\begin{eqnarray}
\nonumber
\present(M,t) \land
\present(M,t') \land
\present(sm,t) \land 
\present(sm,t') \land \\
\present(lm,t) \land 
\neg \present(lm,t') \land
\neg \present(lm',t) \land
\present(lm',t')
\end{eqnarray}

\smallskip
The relational claims are simple: first the two legal concepts are different; second if the social relationship is classified by the social marriage concept at a time, then it has to satisfy the legal concept existing at that very time.
\begin{eqnarray}\label{eq:marriage1}
lm \neq lm' \land 
\classifies(sm,M,t) \to \classifies(lm,M,t) \land 
\classifies(sm,M,t') \to \classifies(lm',M,t')
\end{eqnarray}

The same concept of social marriage ($sm$) persists through time, from $t$ to $t'$ while changing its legal characterization (from $lm$ to $lm'$). For $sm$ to classify a marriage relationship $M$ at $t$, it is necessary that $M$ is also classified as a legal marriage $lm$ (so satisfying concept $lm$ is necessary at $t$ for $sm$), while at $t'$ it is necessary that $M$ is classified by $sm$ which now depends on $lm'$.  

The model presented here is quite natural in {\dolce} for this kind of scenarios. By changing the assumptions we made in the initial discussion of this case, other approaches can be put forward like, e.g., the use of role theory applied to concepts. 
Note also that these modeling approaches are not limited to purely social concepts. They apply to technology-dependent concepts like, e.g., that of road which has different qualifications across history (e.g. in ancient Rome, during the 19th century or today).

%%%%%%%%%%%%%%%%%%%%%%%%%%%%%%
\section{Ontology usage and community impact} \label{sec:usage}
%%%%%%%%%%%%%%%%%%%%%%%%%%%%%%

Foundational ontologies enjoy a double-edged reputation in several communities, spanning across conceptual modeling, semantic web, natural language processing, etc. They are intuitively needed by most data-intensive applications, but their precise utility at different steps of design methodologies is not widely agreed, and certainly not for the same reasons.
As a consequence, the wide application of \dolce ranges from the simple reuse of a few categories, to delving into full-fledged axiomatic versions.
We provide here a quick description of the OWL version of \dolce, a list of application areas and specific reuse cases, with a few comments on the current opportunity for foundational approaches to ontology design.
(For the new CLIF and OWL versions of \dolce produced for the ISO 21838 standard under development, see \url{http://www.loa.istc.cnr.it/index.php/dolce/}).

\dolce ``lite'' versions take into account the requirements from semantic web modeling practices, and the need for simplified semantics as in natural language processing lexicons. They also address the need for some extensions of \dolce categories, by reusing the \dns (Description and Situations) ontology pattern framework, which was early designed to overcome the expressivity limits of OWL, later much facilitated by \textit{punning} in OWL2 \cite{w3c2012owl2} (i.e. the ability to use a constant as the name for a class, an individual, or a binary relation).

In particular, the DOLCE+\dns Ultralite\footnote{\url{http://www.ontologydesignpatterns.org/ont/DUL/dul.owl}} (\dul) OWL ontology was intended to popularize \dolce to the Semantic Web community. \dul uses \dolce, \dns, and a few more ontology design patterns (Plan\footnote{\url{http://www.ontologydesignpatterns.org/cp/owl/basicplan.owl}}, Information Object\footnote{\url{http://www.ontologydesignpatterns.org/cp/owl/informationrealization.owl}}, and Collection, that extend \dolce.  \cite{presutti2016dolce} give an account of \dul as an architecture of ontology design patterns inspired by those integrated theories, and \cite{DBLP:journals/aamas/Gangemi08} offers an integrated axiomatization of plans, information objects and collections in \dns. \dul is the result of various refinements and integrations of the OWL versions of those theories. The main motivations why \dul was conceived include: (i) intuitive terminology (e.g. substituting Endurant and Perdurant with Object and Event), (ii) lighter axiomatization (e.g. giving up some predicate indexing), (iii) integration of other theories, (iv) semantic-web-oriented OWL2 modeling styles.

As reported in \citep{presutti2016dolce}, even a non-exhaustive search makes one stumble upon the great variety of \dul reuse, citing 25 large ontology projects for: e-learning systems, water quality systems; in multimedia: annotation facets, content annotation, audiovisual formal descriptions; in medicine: for modelling intracranial aneurysms, annotating medical images and neuroimages, and for modelling biomedical research; law; events; geo-spatial data; robotics and automation; industry and smart products, textile manufacturing; cybersecurity; enterprise integration; process mining; disaster management; semantic sensor networks; customer relationship management.

In addition, \dul has been applied as a \textit{tool} to improve existing semantic resources. This has happened for example in identifying and fixing millions of inconsistencies in DBpedia, on-the-go discovering modelling anti-patterns that were completely opaque to the axioms of the DBpedia ontology~\citep{DBLP:conf/semweb/PaulheimG15}. Another example is the \dul application to improve the quality of lexical resources, from the very inception of \dolce, used to reorganize the WordNet top level and causing Princeton WordNet developers to include the individual/class distinction in their lexicon \citep{DBLP:journals/aim/GangemiGMO03}, to the recent massive Framester knowledge graph \citep{GangemiAAPR16}, which unifies many different linguistic databases under a frame semantics, and maps them to widely used ontologies under a common \dul hat. Several other standard or {\em de facto} standard are based on or compatible with \dul, e.g., CIDOC CRM (CIDOC Conceptual Reference Model)\footnote{\url{http://www.cidoc-crm.org/}}, SSN (Semantic Sensor Network Ontology)\footnote{\url{https://w3c.github.io/sdw/ssn}} and SAREF (Smart REFerence Ontology)\footnote{\url{https://saref.etsi.org}}.

\medskip
An important lesson learnt is that \dolce can be used to foster different design approaches: 
\begin{enumerate}
\item as an \textit{upper ontology}, in order to support a minimal agreement about a few distinctions;
\item as an \textit{expressive axiomatic theory}, in order to associate one's ontological commitment to well-defined criteria, and to perform (detailed) meaning negotiation;
\item as a coherence/consistency \textit{stabilizer}, able to reveal problems in a conceptualization against both its domain schema, and the data. This approach could also be used to reveal \textit{unwanted} inferences, even when no inconsistency emerges;
\item as a source of \textit{patterns} that improve the quality of ontologies by applying the good practices encoded in \dolce, and eventually ameliorating semantic interoperability.
\end{enumerate}

Especially (3) and (4) are central to the current needs of the huge knowledge graphs maintained by the Web stakeholders, but also (2) is finally emerging as a potential tool to help clarifying the underlying semantics in domains that have been less prone to formalization in the past (e.g. sociology).

%%%%%%%%%%%%%%%%%%%%%%%%%%%%%%
\section*{Acknowledgements} 

Over the years many people have contributed to the application of {\dolce}, to the discussion of the best modeling approaches, and to the development of {\dolce}'s modular extensions. We take the opportunity to thank in particular Emanuele Bottazzi, Francesco Compagno and Alessandro Oltramari. 
{\dolce} was conceptualized and developed as part of the European project WonderWeb (IST-2001-33052) and many public and industrial projects reused it since then. Among these, the authors thank the European project OntoCommons (GA 958371) for co-funding the writing of this paper.

%\nocite{*}
\bibliographystyle{ios2-nameyear}  % Style BST file.
\bibliography{biblioDolceHand,bibliography}

\end{document}